\pdfoutput=1
\documentclass[letterpaper, 10 pt, conference]{ieeeconf}  

\IEEEoverridecommandlockouts                              

\title{\LARGE \bf
When Rolling Gets Weird: A Curved-Link Tensegrity Robot for Non-Intuitive Behavior}

\author{Lauren Ervin$^{1}$, Harish Bezawada$^{1}$, and Vishesh Vikas$^{1}$
\thanks{*This work was supported in part by USDA/NIFA Award {\#2023-67022-40918}. The material contained in this document is based upon work supported in part by a National Aeronautics and Space Administration (NASA) grant or cooperative agreement. Any opinions, findings, conclusions, or recommendations expressed in this material are those of the authors and do not necessarily reflect the views of NASA. This work was supported through a NASA grant awarded to the Alabama/NASA Space Grant Consortium.}
\thanks{$^{1}$Lauren Ervin, Harish Bezawada, and Vishesh Vikas are with the Agile Robotics Lab, University of Alabama, Tuscaloosa, AL 35487, USA
        {\tt\small \{lefaris, hbezawada\}@crimson.ua.edu, vvikas@ua.edu}}%
}

\usepackage{graphics} 
\usepackage{epsfig} 
\usepackage{mathptmx} 
\usepackage{times} 
\usepackage{amsmath} 
\usepackage{amssymb}  
\usepackage{bm,soul,xcolor}
\usepackage{cite}
\usepackage{tikz}
\usetikzlibrary{arrows.meta, automata, positioning, quotes}
\usepackage{comment}
\usepackage[linesnumbered,ruled,vlined]{algorithm2e}
\usepackage{graphicx}
\usepackage{subcaption}
\usepackage{caption}
\usepackage{mathtools}
\usepackage{gensymb}
\usepackage[hidelinks]{hyperref}
\usepackage{stfloats}
\let\labelindent\relax 
\usepackage{enumitem}
\graphicspath{{Figures/}}
\newcommand{\Fig}{Fig. }

\newcommand{\bigse}{{SE}(3)}

\usepackage{amssymb}
\renewcommand{\Re}{\mathbb{R}}

\usepackage[export]{adjustbox}
\begin{document}

\maketitle

\begin{abstract}
Conventional mobile tensegrity robots constructed with straight links offer mobility at the cost of locomotion speed. 
While spherical robots provide highly effective rolling behavior, they often lack the stability required for navigating unstructured terrain common in many space exploration environments. 
This research presents a solution with a semi-circular, curved-link tensegrity robot that strikes a balance between efficient rolling locomotion and controlled stability, enabled by discontinuities present at the arc endpoints. 
Building upon an existing geometric static modeling framework \cite{TeXploR2024}, this work presents the system design of an improved Tensegrity eXploratory Robot 2 (TeXploR2). 
Internal shifting masses instantaneously roll along each curved-link, dynamically altering the two points of contact with the ground plane. 
Simulations of quasistatic, piecewise continuous locomotion sequences reveal new insights into the positional displacement between inertial and body frames. 
Non-intuitive rolling behaviors are identified and experimentally validated using a tetherless prototype, demonstrating successful dynamic locomotion. 
A preliminary impact test highlights the tensegrity structure's inherent shock absorption capabilities and conformability.
Future work will focus on finalizing a dynamic model that is experimentally validated with extended testing in real-world environments as well as further refinement of the prototype to incorporate additional curved-links and subsequent ground contact points for increased controllability.
\end{abstract}
\section{INTRODUCTION} 
\begin{figure}[!h]
    \centering
    \includegraphics[width = 0.95\linewidth]{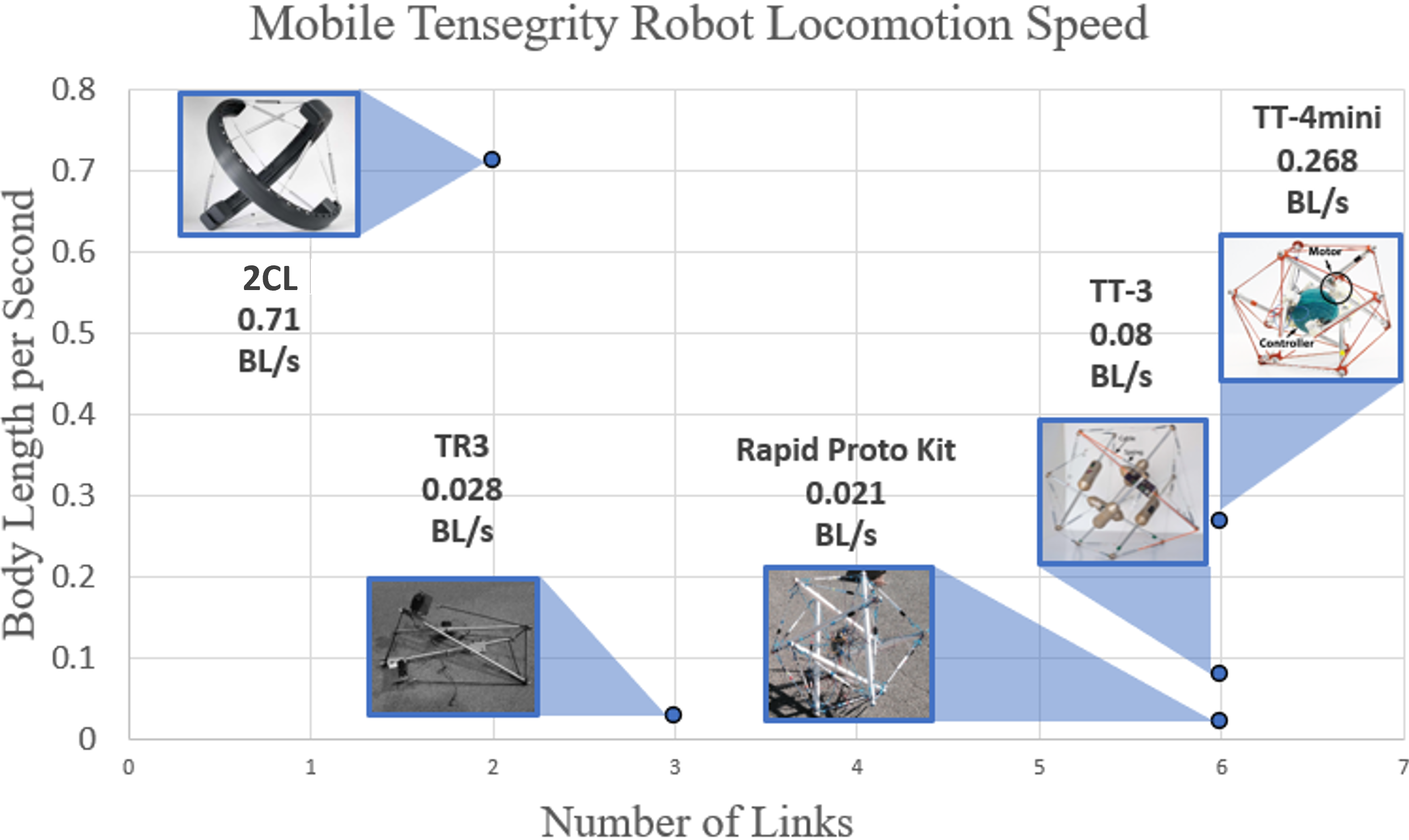}
    \caption{Tensegrity robot locomotion speed comparison among five different mobile tensegrity robots with published experimental speeds \cite{bohm_etal_mechatronics_2016, paul_etal_ieee_transactions_robotics_2006, chen_etal_mechanisms_robotics_2016, kim_rapid_2014, chen_inclined_2017}. The four straight-link designs are significantly slower than the curved-link design when normalized to body lengths per second (BL/s).}
    \label{fig:comparison}
\end{figure}

Much of the motivation for adopting semi-circular, rigid links as the structural foundation of the proposed tensegrity structure lies in the goal of enhancing locomotion speed of tensegrity systems. Improved speed is particularly beneficial for increasing potential data acquisition efficiency during science missions in certain unstructured environments of interest such as the lunar surface. Traditionally, the vast majority of mobile tensegrity robots utilize rigid, straight links arranged in icosahedrons \cite{7139590, ken_2014, pietila_etal_aiaa_2011, hirai_etal_robio_2012, paul_etal_ieee_transactions_robotics_2006, chen_etal_mechanisms_robotics_2016, kim_rapid_2014, chen_inclined_2017}. These designs typically rely on active shape morphing through cable lengthening and shortening to achieve movement. While incorporating multiple actuators can enhance controllability, this actuation strategy introduces significant coordination complexity. The need for precise synchronization among numerous actuators often leads to increased latency in movement sequences where the actuators must coordinate with one another in unison. If cable length changing does not occur simultaneously, it will add even more time to movement sequences. The combination of the straight, rigid bars held together with changing cable lengths generally produces toppling-like gaits, further slowing down planar locomotion compared to smooth rolling. An example of this slowed behavior is shown in \Fig~\ref{fig:comparison}. The four prototypes designed with straight links \cite{paul_etal_ieee_transactions_robotics_2006, chen_etal_mechanisms_robotics_2016, kim_rapid_2014, chen_inclined_2017} have reported travel speeds ranging from 0.021 and 0.268 body lengths per second (BL/s). In contrast, the prototype designed with curved-links \cite{bohm_etal_mechatronics_2016} achieved a speed of up to 0.71 BL/s, or nearly three times faster than the fastest straight-link design. 

Speed is not the only critical factor. Otherwise, a wheeled or spherical design would be preferable; spherical robots with internal mass movement can exhibit incredibly efficient and smooth rolling behavior \cite{8977011, 6766197}. Although spherical robots excel at rolling speed, they lack additional points of contact with the ground plane to quickly stop. Innovative designs that strike a balance between speed and adaptation to non-uniform terrains are attractive for those in the space, water,  and search and rescue communities. A curved-link tensegrity design can yield efficient rolling behavior while introducing stability at arc endpoints that can handle inclines. Additional conformability and compliance inherent to the structure can tolerate sharp edges and uneven surfaces.

Regardless of the geometry of the rigid elements used, modeling the connections between members and cables in tensegrity systems remains an open challenge and understudied area of research due to their non-linear characteristics. Although dynamic modeling is still an ongoing effort within the field, the geometric nature of these systems lends itself well to advanced mathematical tools such as Screw Theory and Lie groups, which offer robustness against singularities \cite{liu_etal_cac_2019, andreas_kin, andreas_dyn}. Building upon a previously developed static modeling framework that leverages geometric representations and Screw Theory \cite{TeXploR2024}, this work introduces an analysis of quasistatic and non-intuitive behaviors. This non-intuitive control enables faster transitioning between locomotion states, further increasing the controllability of the system. 

\begin{figure}[!ht]
    \centering
    \includegraphics[width = 0.95\columnwidth]{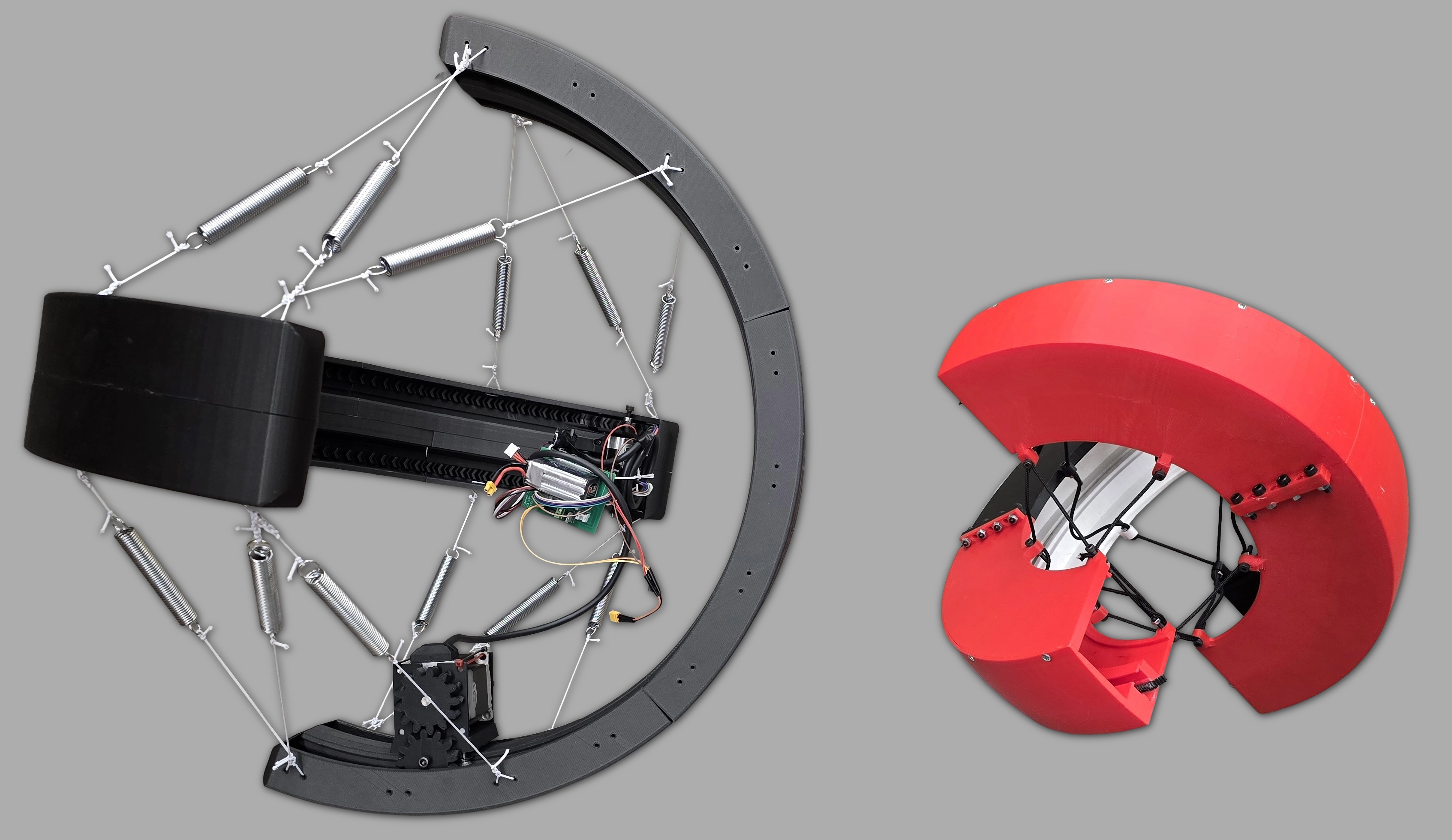}
    \caption{TeXploR prototype on the right (red) and TeXploR2 on the left (black). The two curved arcs in each iteration are held together with a network of 12 elastic cables. Rolling is achieved via internal mass shifting along the arcs, which shifts the overall CoM of TeXploR2. TeXploR2 is signficantly larger and weighs nearly triple that of TeXploR.}
    \label{fig:v1v2}
\end{figure}

The latest iteration of the TeXploR prototype is presented on the left in \Fig~\ref{fig:v1v2}. This design is a newer version than the previous tetherless prototype \cite{TeXploR2024} shown on the right and takes inspiration from \cite{bohm_etal_mechatronics_2016} and \cite{gim_ringbot_2024}. The new robot demonstrates smoother movement capable of dynamic rolling, and a larger open central area is included to accommodate future additions such as suspended gimbals and science payloads.

\textbf{Contributions.}
This work presents a design methodology for a Tensegrity eXploratory Robot 2 (TeXploR2) capable of dynamic movement. Simulations show a quasistatic rolling sequence where TeXploR2 switches between four states in a piecewise continuous path. Non-intuitive behavior for switching states is analyzed, and rolling experiments with the tetherless prototype show successful validation. Vicon motion capturing data of the experiments further highlight the added impact test to show the resilience of the system and the inherently compliant tensegrity design.

\textbf{Paper Organization.}
The next section discusses the system design of the improved, tetherless TeXploR2 prototype along with design requirements. The third section briefly visits the geometric relationship between the two arcs and shows quasistatic simulation for a full rolling sequence between four states of locomotion.  Section four shows a dynamic rolling experiment with the physical prototype that utilizes non-intuitive behavior for more efficient state jumping and an impact test.  Finally, the Conclusions section discusses the findings in this work as well as next steps for the research.

\section{System Design}\label{TeXDesign}


This version represents an improvement over the previous tetherless prototype depicted in \Fig~\ref{fig:v1v2}. Each shifting mass, $m_i$, in the updated design weighs 1,150g which is nearly three times heavier than that of the earlier prototype. Similarly, the semi-circular curved arc weighs approximately 1,300g, which is slightly more than triple the mass of the previous arcs. While the current shifting mass to arc mass ratio is approximately 0.88:1, this can easily be adjusted to reach a 1:1 ratio or beyond by adding supplemental mass as needed. However, based on experimental testing, the 0.88:1 ratio was found to be sufficient. The new prototype also features larger dimensions: the arc diameter has increased from 403mm to 585mm, and similarly the width has grown from 83mm to 98mm. To improve durability and flexural strength, the arcs are fabricated using Onyx\textsuperscript{TM} (from Markforged) via additive manufacturing, replacing the tough PLA used in the previous design. For future deployments in extreme environments such as the arctic, alternative materials may be required to ensure structural compliance and performance reliability.

\subsection{Design Requirements}

Following the development and evaluation of the initial prototype, several improvements were identified and incorporated into the updated design. Based off those lessons learned, a refined set of design requirements for the second generation TeXploR2 are as follows.

\begin{enumerate}[leftmargin=*]
    \item \textit{Shifting mass smoothness:} Regardless of the orientation of the two masses, they must maintain continuous contact and proper alignment along the track throughout dynamic motion. This includes minimizing any rotation, rocking, or gaps between the shifting mass assembly and the arc track during dynamic movement. Preliminary manual alignment tests have demonstrated consistent and smooth rolling behavior along the arc regardless of shifting mass location. However, to further quantify this, an onboard IMU will measure the acceleration in future experiments to ensure there are no sudden unintended accelerations or decelerations.
    \item \textit{Arc symmetry:} Both the curved arcs and the integrated gear racks must be designed to prioritize geometric and mass symmetry. This ensures consistent center of mass movement, regardless of which side of the arc faces the ground. This is intended to reduce potential discrepancies between the arc mass in the model and the mass distribution along the physical arc in the prototype. This is opposed to the initial design, where mass distribution along an arc was not evenly distributed along the circumference of the arc, leading to discrepancies between modeled and actual behavior. 
    \item \textit{Scale:} The structure must reserve at least $10cm^3$ of open space in the center between the two arcs to allow for future integration of additional payloads, e.g., a suspended vision system or science payload.   
    \item \textit{Modular and adaptable:} The overall design should support modular expansion and adaptability; it should be invariant to the specific primitive. Specifically, it must accommodate an increased number of curved links without requiring a fundamental redesign. Adding more curved links will expand the number of ground contact points, thereby increasing the number of hybrid system states and improving system controllability. Additionally, the arc-to-arc connection points (nodes) are also modular. Although only eight node positions are required to complete the cable routing, each arc includes a total of 48 holes, providing flexibility for future reconfiguration and adjustments as needed.
\end{enumerate}

\subsection{Motor Carriage Design}

\begin{figure}[!ht]
    \centering
    \includegraphics[width = 0.95\columnwidth]{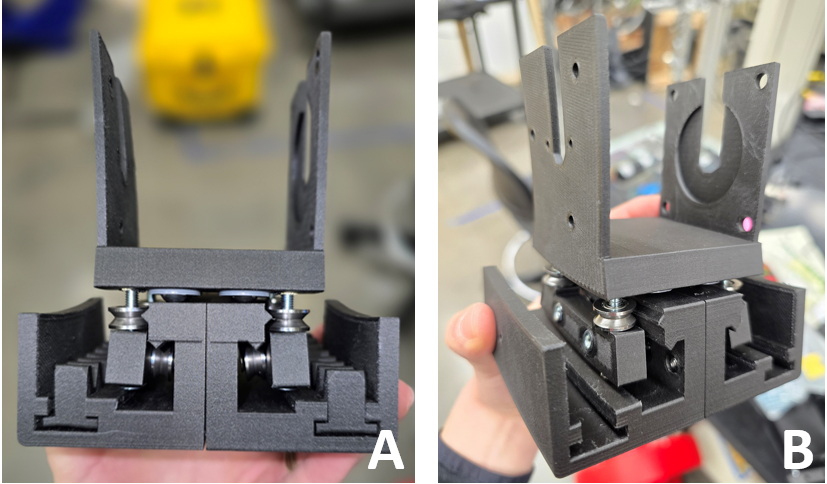}
    \caption{Motor carriage design. a) Cross-section view of the v-groove and ball bearing configuration in the motor carriage that grip the T-track. b) Isometric view of the motor carriage attached to the T-track. This view emphasizes the v-groove bearing attachments at both ends of the motor carriage, preventing motor rocking during increased velocity.}
    \label{fig:track}
\end{figure}

\begin{figure}[!ht]
    \centering
    \includegraphics[width = 0.95\columnwidth]{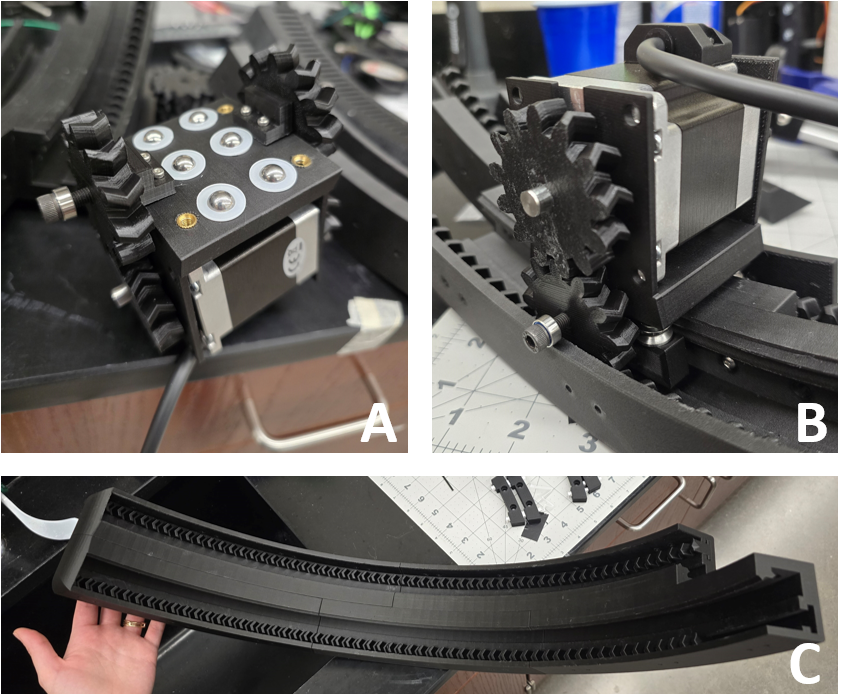}
    \caption{Curved arc with motor assembly. a) Motor carriage underside with matching arc curvature for smooth rolling along the top of the T-track. Six embedded ball bearings promote smoothness. Pinion gears are connected via bolts anchored into heat set inserts on the underside and sides of the carriage. b) The motor carriage fixed to the curved arc. c) Internal geometry of the sister gear racks. Partial assembly highlights the connection points.}
    \label{fig:carriage}
\end{figure}

As previously described, actuation of the system is achieved through controlled movement of a shifting mass along each curved arc. This movement is facilitated by a dual-shaft NEMA23 stepper motor assembly that traverses each arc in a coordinated manner. To maintain proper tension and alignment across all potential configurations, several critical design choices were implemented. The motor carriage, shown in \Fig~\ref{fig:track}, latches onto an internal T-track that defines the inner curvature of each arc. To guide the carriage and minimize unwanted motion, the sides and underside of the top of the track include four strategically placed triangular protrusions. The protrusion geometry matches eight v-groove bearings, i.e. four oriented horizontally and four oriented vertically. The horizontal bearings mitigate torsional rotation that might otherwise occur from the motor's tilted mounting angle of approximately $45^{\circ}$ relative the ground plane movement due to the motor laying at an approximate $45^{\circ}$ from the ground plane. Meanwhile, the vertical bearings secure the carriage when hanging upside down along the arc. Six embedded ball bearings on the underside of the carriage contribute to smooth rolling along the top of the T-track with a matching arc geometry, as illustrated in \Fig~\ref{fig:carriage}. 

\subsection{Arc Design}


Each semi-circular arc contains two internal gear racks positioned parallel to each other along the arc's inner curvature. A dual-shaft NEMA23 stepper motor drives matching pinion gears that engage simultaneously with both gear racks on a single arc.  Both the gear racks and pinions feature a herringbone teeth design that was deliberately selected to increase the contact ratio and encourage smooth power transmission, even at high velocity or under significant loads. 

The symmetrical design ensures consistent mass distribution from the arcs, regardless of the robot's orientation. This consistency helps minimize discrepancies between the modeled behavior and physical prototype. The width of each arc is tightly coupled to the stepper motor's dimensions; the stepper motor width dictates the arc's minimum width. For example, the NEMA23 stepper motor, mounted at the center of the arc, measures 56mm in width. An additional 13mm on each side accounts for the pinion gears fixed to the motor's dual shafts. The arc's outer walls are 5mm wide, and a 3mm clearance is maintained between the inner wall of the arc and the gear rack to prevent unintended friction. In the prototype presented, these constraints result in a minimum viable arc width of 98mm to preserve symmetry. Although the arc width could be expanded beyond this and adapted into various shapes, maintaining a balanced weight distribution is essential; a higher ratio of shifting mass to arc mass allows for reduced displacement of the shifting mass to achieve rolling motion. In other words, lighter arcs enhance power efficiency. Additionally, the arcs are designed to be relatively flat rather than fully curved, as shown in \Fig~\ref{fig:track}. This geometry promotes rolling along a defined edge that is hypothesized to reduce discrepancies with the model that assumes a single point per ground contact point. More prototypes that alter the defined edge thickness as a control parameter could further validate this idea. The model can also account for arc width during state transitions, i.e. changes in the pivot point.

\subsection{Electronics}
All electronics onboard TeXploR2 interface with a Raspberry Pi Zero 2W and a custom PCB. The system actuators are dual shaft NEMA23 bipolar stepper motors capable of exerting up to 1.85Nm of torque and rotating 200 steps per revolution in full-step mode. Six full rotations are required to travel one semi-circular arc. The holding torque of one motor is significantly higher than what is required to shift the mass of the system; this ensures payloads can be added in the future without the need for swapping in stronger actuators.  They are each controlled via high-current TB67S128FTG motor driver carrier boards. These drivers support Active Gain Control (AGC) which automatically reduces the current when the maximum torque is not required, further reducing power consumption. Microstepping is also supported to enable finer resolution in motor control. Each shifting mass is equipped with a 9-axis BNO085 IMU capable of capturing both orientation and acceleration data that is communicated via I2C at a rate of 200Hz. Most electronic components are mounted on a single motor assembly, with the exception of one IMU and two limit switches. TeXploR2 is operated remotely by establishing an SSH connection to the Pi via a shared network. Through this interface, various commands are issued to control the two stepper motors. Currently, the system uses open loop control with a set of predefined rolling sequences in which only one motor actuates at a time. The entire system is powered by a single 11.4V 1.5Ah LiPo battery, enabling tetherless capabilities.

\begin{figure}[htbp]
    \centering
    \includegraphics[width=0.95\columnwidth]{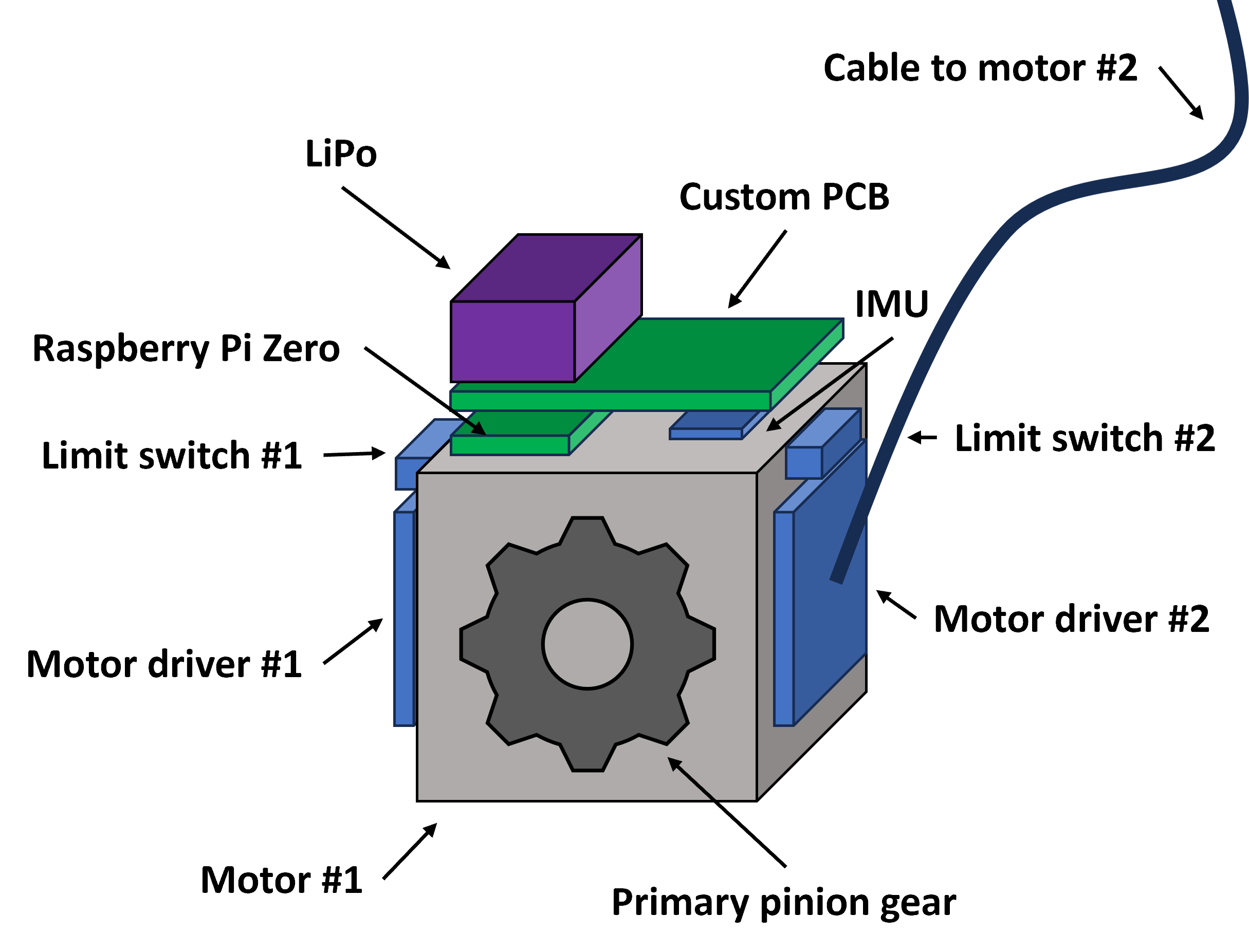}
     \caption{TeXploR2 electronics. All electronics are rigidly connected to an electronics mount that sits on top of a motor.}
    \label{fig:pcb}
\end{figure}

\subsection{Shock Absorption}
\begin{figure}[!h]
    \centering
    \includegraphics[width = 0.95\columnwidth]{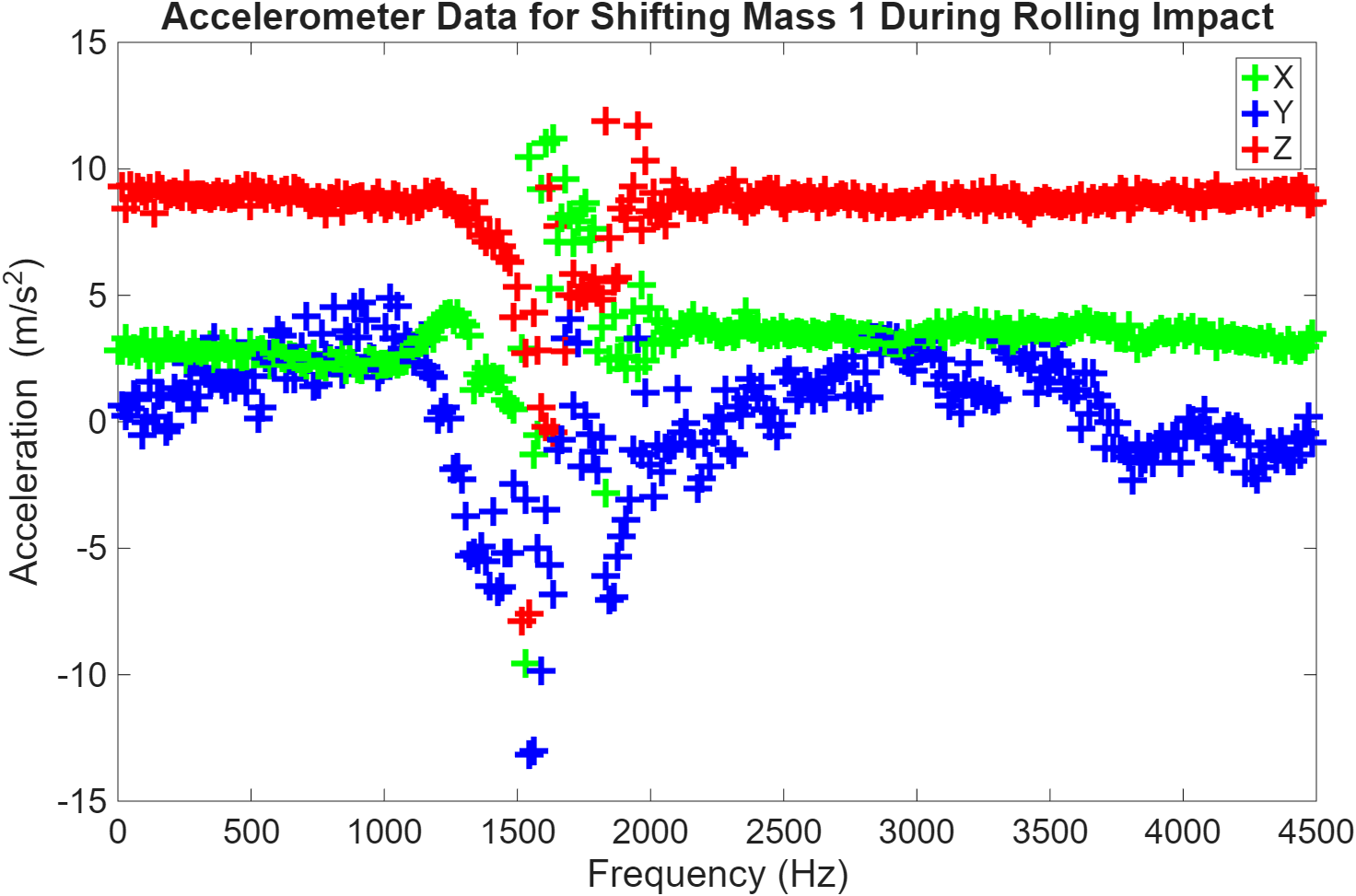}
    \caption{BNO085 acceleration collected during an impact. It is absorbed within 5 seconds and does not impact mass shifting.}
    \label{fig:accelimpact}
\end{figure}

Due to the compliance of the system, TeXploR2 has a level of shock absorption built into the mechanical structure. The compressed members are connected with a network of tensioned cables and springs that absorb collisions or drops with minimal impact to the system. \Fig~\ref{fig:accelimpact} highlights acceleration readings from an accelerometer connected to one of the shifting masses while TeXploR2 drops off of a platform during a rolling sequence. When TeXploR2 hits the ground, there are oscillations from the shock absorption that subside after roughly 5 seconds. Notably, the driving shifting mass continues to move during this recovery time.

\subsection{Further Design Improvements}

The arc design can, and should, be further adapted for specific surface traversal scenarios and specialized applications. For instance, when operating on granular terrains such as sand or lunar regolith, it is necessary to incorporate grousers into the outer surface of the arcs. This feature will help prevent excessive sinking by concentrating the robot's mass onto discrete contact points, rather than distributing it across the full arc. As a result, localized penetration into the granular surface may occur without necessarily submerging the whole arc into the medium. Corresponding adjustments to the robot model would be required to accurately reflect these changes, which would likely shrink the number of stable configurations available to TeXploR2. In scenarios involving steep inclines, the integration of directional or non-directional microspines along the arc surface could enhance traction and grip stability. These microspines passively latch onto surface asperities, potentially increasing TeXploR2's ability to navigate irregular or non-uniform terrain. Both grousers and microspines represent forms of embodied intelligence and offer promising avenues for expanding TeXploR2's capabilities in unstructured and/or extreme environments.

Future enhancements to the electronics of TeXploR2 may be added to expand the system's capabilities.  While most tensegrity robots achieve active shape morphing through the alteration of cable segment lengths, TeXploR2 currently relies on internal mass shifting.  To incorporate active shape morphing into the existing design, linear actuators would need to be integrated into each of the twelve connecting cable segments. Although this would add complexity into the system, it would also provide greater controllability and maneuverability. Notably, the current design already supports passive shape morphing. By manually detaching the twelve connecting cable segments, the structure can collapse from a spatial configuration to a planar one, reducing its size to approximately 1/6th of its deployed volume. This transformation significantly decreases the system's physical footprint and associated transportation expenses, particularly for costly space deployment scenarios. Additionally, scaling up the system, i.e. increasing the arc size, creates more open volume at the robot's center when fully deployed. This space could be utilized in future iterations to suspend payloads such as sensors for obstacle avoidance, path planning, SLAM, etc. However, such upgrades would introduce new challenges, particularly in accurately estimating the transformation matrix between the sensor payload, robot, and the global inertial frames. Specifically, projecting sensor data into a global frame will require precise tracking of a suspended payload's motion relative to both the robot's center of mass as well as reference frames fixed to the two arcs. Future implementations will incorporate closed loop feedback control for more precise and adaptive navigation. Addressing these challenges will be crucial in incorporating closed loop feedback control for more precise and adaptive navigation and advanced perception in future versions of TeXploR2.
\section{QUASISTATIC ANALYSIS}

\begin{figure}[!ht]
    \vspace{0.5em}
    \centering
    \includegraphics[width = 0.95\columnwidth]{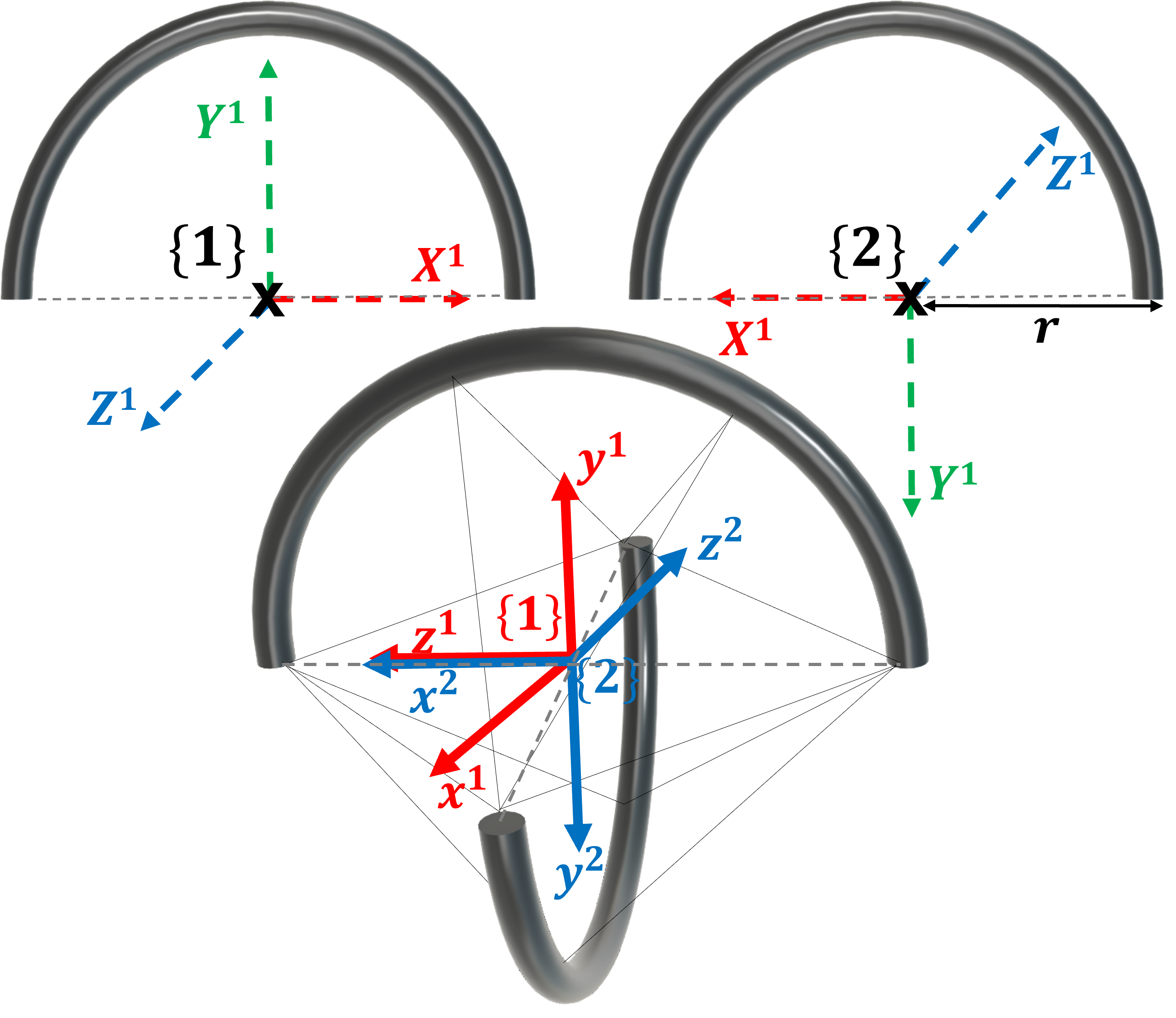}
    \caption{Visualization of the relationship between arcs $L_1$ and $L_2$. The robot body $\{b\}$ coordinate frame coincides with $\{1\}$.}
    \label{fig:frames}
\end{figure}

The kinematics of TeXploR2 closely match the static modeling framework presented in the previous work \cite{TeXploR2024}. The two semi-circular arcs $L_1,L_2$ are represented with radii $r=272.5mm$ and masses $m_1,m_2=1,300g$. The origin of the two coordinate systems $\{1\},\{2\}$ are fixed at the base of the center of the arc, such that the $x$ axis is parallel to the line running through the arc endpoints and the $z$ axis is normal to the arc plane as shown in \Fig~\ref{fig:frames}. 

Each motor consists of mass $M_i=1,150g$ that travels along one of the arcs. At any instance in time, the moving mass's location about the arc is given by $p_i$ and the angle at which it has traveled relative to the center of the arc is $\theta_i$. The instantaneous point of contact the arc makes with the ground plane is represented by $q_i$. Similarly, the angle of the point of contact relative to the center of the arc is given by $\phi_i$. Consequently, the points $p_i,q_i$ can be represented as

\begin{equation}
\begin{gathered}
\bm{p}_1^b = r\begin{bmatrix}
        \cos(\theta_1)\\ \sin(\theta_1)\\0
    \end{bmatrix}, \quad
\bm{p}_2^b= \bm{o}_{12} + R_{12}\bm{p}_2^2 = 
        r \begin{bmatrix} 0\\
-\sin(\theta_2)\\
\cos(\theta_2) \end{bmatrix} \\
\bm{q}_1^b = r\begin{bmatrix}
        \cos(\phi_1)\\\sin(\phi_1)\\0
    \end{bmatrix}, \quad 
\bm{q}_2^b= \bm{o}_{12} + R_{12}\bm{q}_2^2 = r\begin{bmatrix} 0\\
-\sin(\phi_2)\\
\cos(\phi_2) \end{bmatrix}
\end{gathered}
\end{equation}

where $R_{12}\in SO(3)$ and $\bm{o}_{12}\in \Re^{3\times 1}$ are the rotation matrix and the displacement vector between the origins of the reference frames $\{1\},\{2\}$. $R_{12}$ and $\bm{o}_{12}$ combine to make the transformation matrix $T_{12}\in \bigse$ that changes the representation of a point from coordinate system $\{2\}$ to $\{1\}$. 

\begin{align}
\begin{gathered}
T_{12}  = \begin{bmatrix}
    R_{12} & \bm{o}_{12}^1\\
    \bm{0}_{1\times 3} & 1
\end{bmatrix}, R_{12} = \begin{bmatrix}
0 & 0 & 1 \\
0 & -1 & 0 \\
1 & 0 & 0 \\
\end{bmatrix}, \bm{o}_{12}^1 =  \begin{bmatrix} 0\\0\\0\end{bmatrix} 
\end{gathered}
\end{align}

\begin{figure}[h]
    \centering
    \includegraphics[width = 0.95\linewidth]{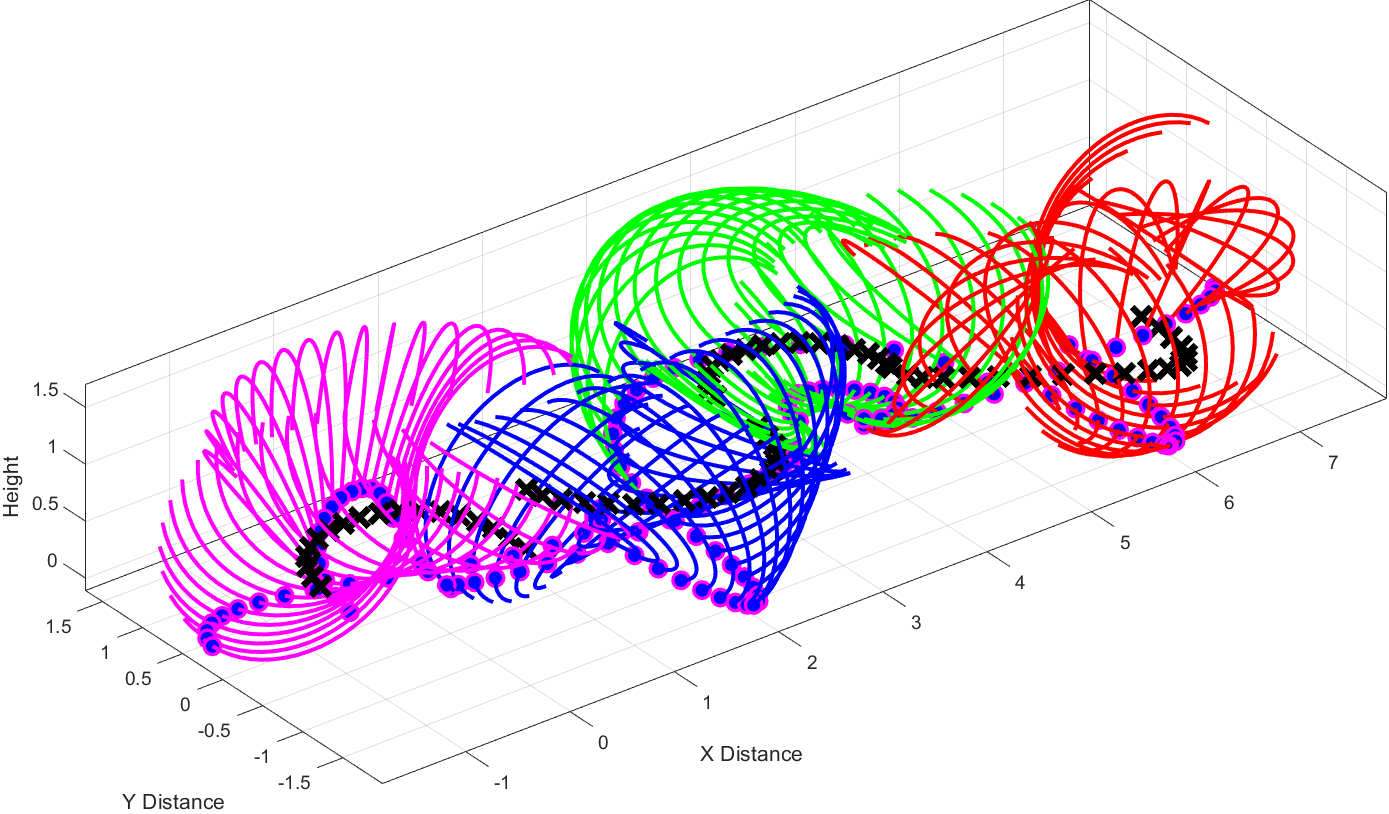}
    \caption{Quasistatic locomotion simulation.  The magenta lines show the movement during state 1, blue lines show the movement during state 4, green lines show the movement during state 2, and red lines show the movement during state 3. The rolling sequence traverses state 1 $\rightarrow$ state 4 $\rightarrow$ state 2 $\rightarrow$ state 3. The dots represent the points of contact, and the black crosses represent the center of mass.}
    \label{fig:fourstates}
\end{figure}

From the static modeling framework, it is defined that TeXploR2 is a hybrid system consisting of four states of locomotion for a rolling sequence. These states are dictated by an arc endpoint contact point acting as a pivot while the other contact point rolls along the ground until it reaches its respective arc endpoint. Instantaneously, this is the new pivot point that stays fixed to the ground plane as the opposing contact point changes due to its internal mass shifting along the entirety of the arc. This continues back and forth with two arcs that each contain two endpoints to result in four states producing a piecewise continuous motion.

\begin{enumerate}
    \item[] \textbf{State 1}: $\phi_1\in (0,180\degree), \phi_2=\phantom{1}0\degree,\phantom{8}$ roll about $L_1$
    \item[] \textbf{State 2}: $\phi_1\in (0,180\degree), \phi_2=180\degree$, roll about $L_1$
    \item[] \textbf{State 3}: $\phi_1=\phantom{1}0\degree,\phantom{8} \phi_2\in (0,180\degree)$, roll about $L_2$ 
    \item[] \textbf{State 4}: $\phi_1=180\degree, \phi_2\in (0,180\degree)$, roll about $L_2$ 
\end{enumerate}

For further details and diagrams of the geometric relationship of the system, we encourage the reader to reference the Robot Kinematics section in the authors' prior work \cite{TeXploR2024}.

\subsection{Simulation}
Quasistatic simulations in \Fig~\ref{fig:fourstates} show TeXploR2 in a full rolling sequence traveling from state 1 $\rightarrow$ state 4 $\rightarrow$ state 2 $\rightarrow$ state 3. Within each state in increments of $10\degree$, the control input of one of the shifting masses, $\theta_i$, increases or decreases to reach either $0\degree$ or $180\degree$ depending on the starting angle while the other control mass, $\theta_{i\pm 1}$, remains stuck to the pivot point ($0\degree$ or $180\degree$ depending on the state).  The robot equilibrium position represented through ground contact angles $(\phi_1,\phi_2)$ is found for each of these input combinations. Additionally, the original static modeling framework generates a closed form solution for $T_{sb}\in \bigse$ which transforms a point from a fixed inertial frame to the body frame using a rotation matrix, $R_{sb}\in SO(3)$, and a translation vector, $\bm{o}_{sb}\in\mathbb{R}^3$. However, the previous static modeling framework assumed that only the $z$ component of $\bm{o}_{sb}$ could be found analytically from $r$ and $x,y$ remained free positions anywhere along the ground plane. For the quasistatic case, it is instead found that $x,y$ components are also dependent on the changing mass positions. Using this information, $\bm{o}_{sb} = [\cos(\theta_i),\sin(\theta_i),\sqrt{r}/2]^T$ is found. Notably, once a ground contact point reaches the end of one of the arcs (equivalently, a transition between the four states), the arc endpoint associated with that robot orientation now acts as a pivot, i.e. that position must be combined with the following displacement vectors. For instance, traveling in a sequence from  state 1 $\rightarrow$ state 4 $\rightarrow$ state 2 $\rightarrow$ state 3 would result in the following displacement vectors.

\begin{enumerate}
    \item[] \textbf{State 1$\rightarrow$4}: $\bm{o}_{sb} = [\cos(\theta_1),\sin(\theta_1),\sqrt{r}/2]^T$ at $\phi_2$
    \item[] \textbf{State 4$\rightarrow$2}: $\bm{o}_{sb} = [\cos(\theta_2),-\sin(\theta_2),\sqrt{r}/2]^T$ at $\phi_1$
    \item[] \textbf{State 2$\rightarrow$3}: $\bm{o}_{sb} = [\cos(\theta_1),\sin(\theta_1),\sqrt{r}/2]^T$ at $\phi_2$
    \item[] \textbf{State 3$\rightarrow$1}: $\bm{o}_{sb} = [\cos(\theta_2),-\sin(\theta_2),\sqrt{r}/2]^T$ at $\phi_1$
\end{enumerate}

The robot configurations for the full sequence rolling between the four states are shown in \Fig~\ref{fig:fourstates}.  Here, the magenta lines show the movement during state 1, green lines show the movement during state 2, red lines show the movement during state 3, and blue lines show the movement during state 4.  The dots represent $q_1, q_2$, and the black crosses show how the center of mass shifts during a rolling sequence.

\section{NON-INTUITIVE DYNAMIC RESULTS}

\begin{figure}[htbp]
    \centering
    \includegraphics[width = 0.95\columnwidth]{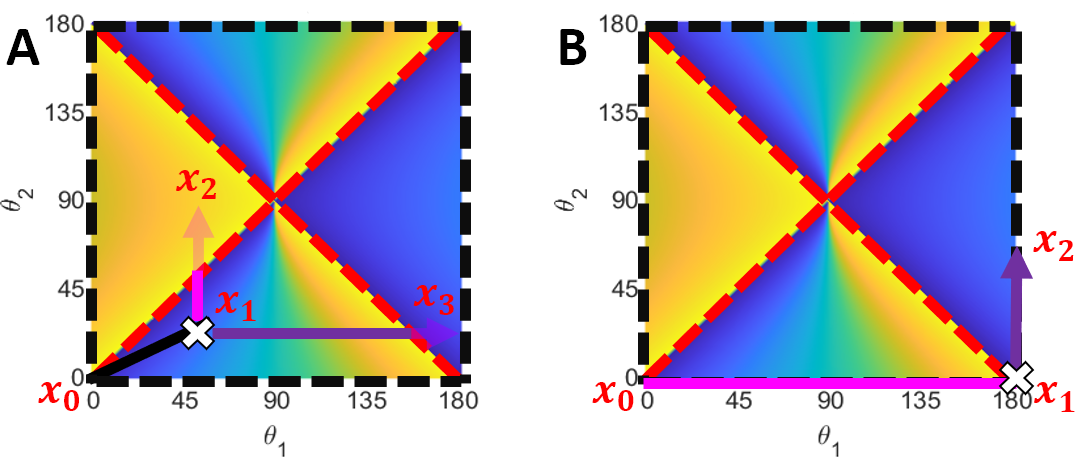}
    \caption{Static equilibrium positions of the four states. a) Two quasistatic control path sequences highlight the state transition boundaries (dotted red line) for non-intuitive behavior: $x_0\rightarrow x_1 \rightarrow x_2$ for state $1 \rightarrow 3$ and $x_0\rightarrow x_1 \rightarrow x_3$ for state $1 \rightarrow 4$. b) An intuitive state transition from $x_0\rightarrow x_1 \rightarrow x_2$ for state $1 \rightarrow 4$ along the black dotted perimeter line.}
    \label{fig:analysis}
\end{figure}

In quasistatic robot configurations, two types of behavior can be exploited when switching states: intuitive and non-intuitive behavior. Non-intuitive behavior occurs when the shifting mass associated with the moving ground contact point does not travel to an arc endpoint before transitioning states. This is shown in two different examples in \Fig~\ref{fig:analysis}a.

Here, the red dotted lines represent the state transition boundaries for the non-intuitive behavior described. A starting position, $x_0$, travels to location $x_1$ where it sits in state 1. It can travel directly upwards, passing the boundary line, to land at $x_2$ in state 3. Similarly, it could instead travel to the right, passing another boundary line, to land at $x_3$ in state 4. The alternative option is to travel along the black perimeter lines as shown in \Fig~\ref{fig:analysis}b. Here, $x_0$ travels the entire state 1 to $x_1$ before entering state 4 at $x_2$. It is clear that traveling along the non-intuitive boundary lines can achieve state jumping more efficiently than the traditional method. In \Fig~\ref{fig:analysis}a, traveling along the non-intuitive path from $x_0\rightarrow x_1 \rightarrow x_3$ for state $1 \rightarrow 4$ would require $\theta_1$ traversing a total of $162.81\degree (50.31\degree \text{ from } x_0\rightarrow x_1, \text{ then } 112.5\degree \text{ from } x_1 \rightarrow x_3)$. Alternatively, the intuitive path in \Fig~\ref{fig:analysis}b from $x_0\rightarrow x_1$ would require $\theta_1$ to travel the full $180\degree$ to transition from state $1 \rightarrow 4$. Next, a dynamic rolling sequence with this non-intuitive behavior is experimentally shown.

\begin{figure*}[!ht]
    \vspace{0.5em}
    \centering
    \begin{subfigure}{0.49\linewidth}
         \centering
         \includegraphics[width=\textwidth]{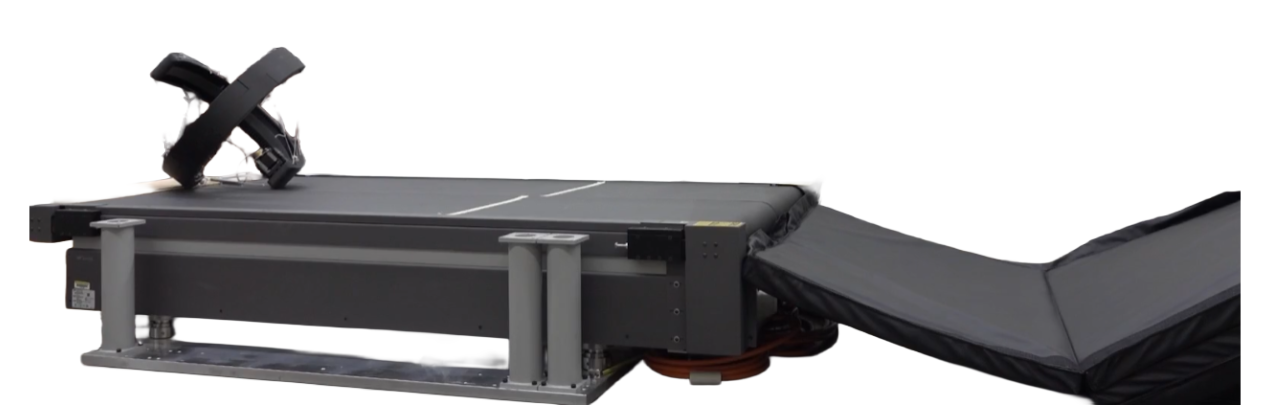}
         \caption{t=1}
         \label{fig:t1}
     \end{subfigure}
     \begin{subfigure}{0.49\linewidth}
         \centering
         \includegraphics[width=\textwidth]{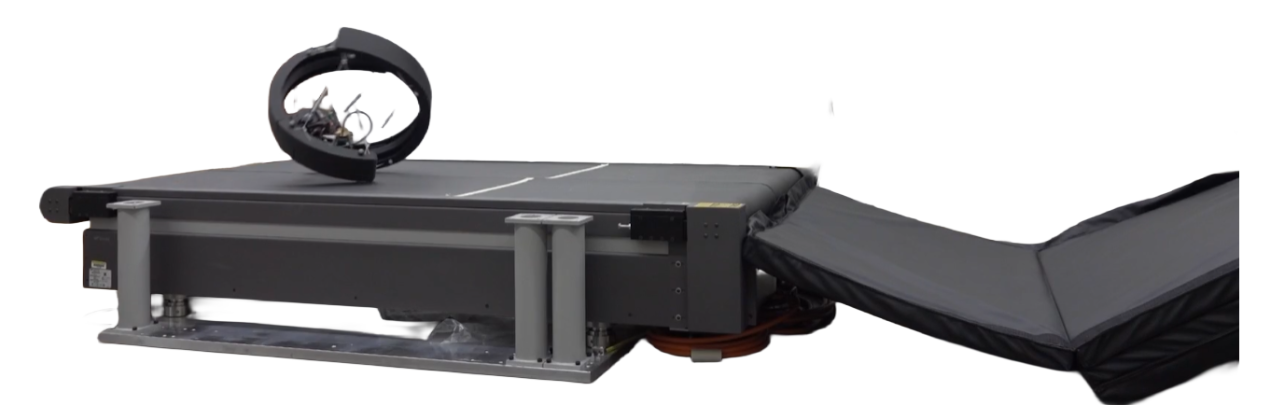}
         \caption{t=3}
         \label{fig:t3}
     \end{subfigure}
     
     \begin{subfigure}{0.49\linewidth}
         \centering
         \includegraphics[width=\textwidth]{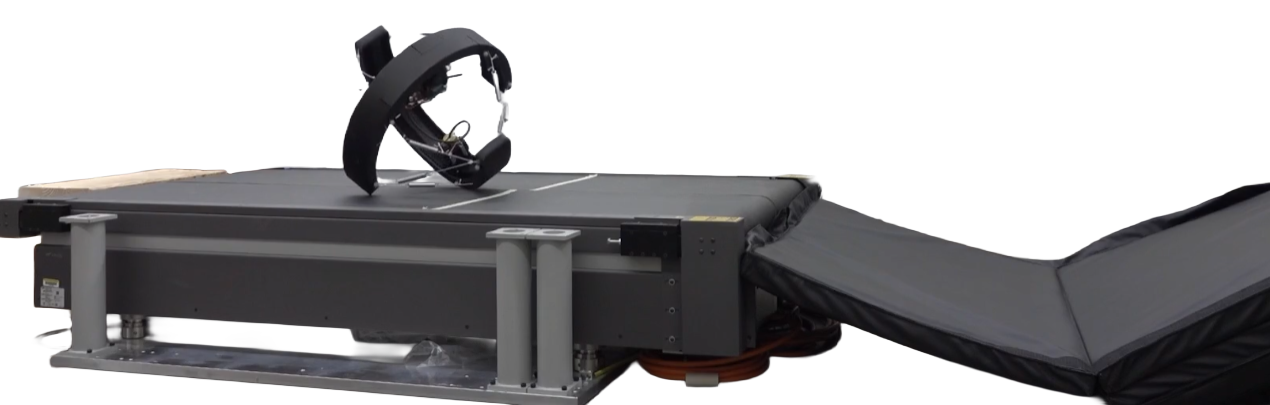}
         \caption{t=5}
         \label{fig:t5}
     \end{subfigure}
     \begin{subfigure}{0.49\linewidth}
         \centering
         \includegraphics[width=\textwidth]{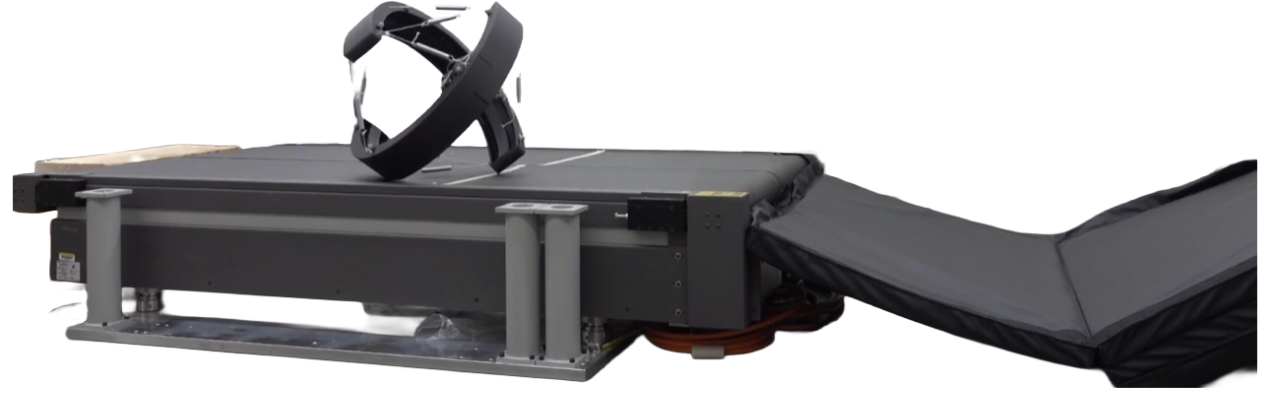}
         \caption{t=7}
         \label{fig:t7}
     \end{subfigure}
     
     \begin{subfigure}{0.49\linewidth}
         \centering
         \includegraphics[width=\textwidth]{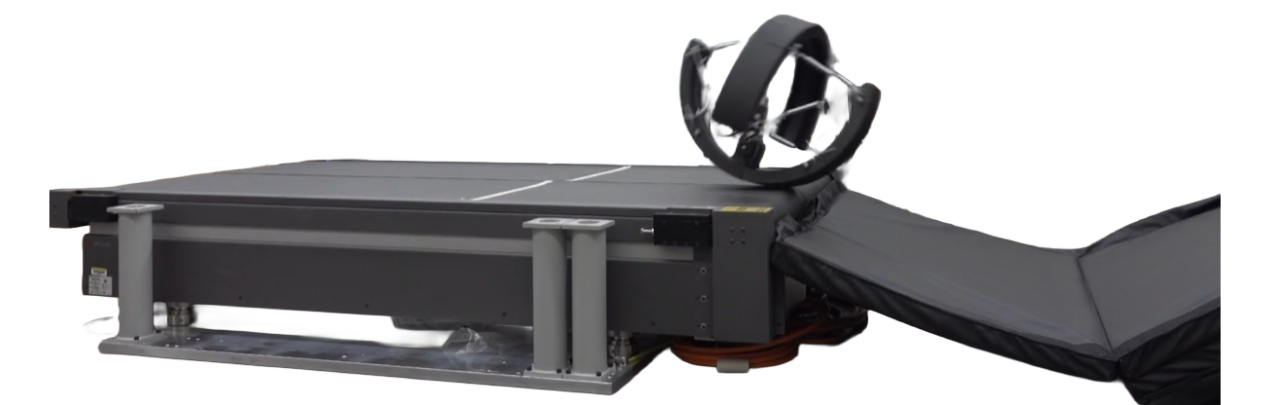}
         \caption{t=8}
         \label{fig:t8}
     \end{subfigure}
     \begin{subfigure}{0.49\linewidth}
         \centering
         \includegraphics[width=\textwidth]{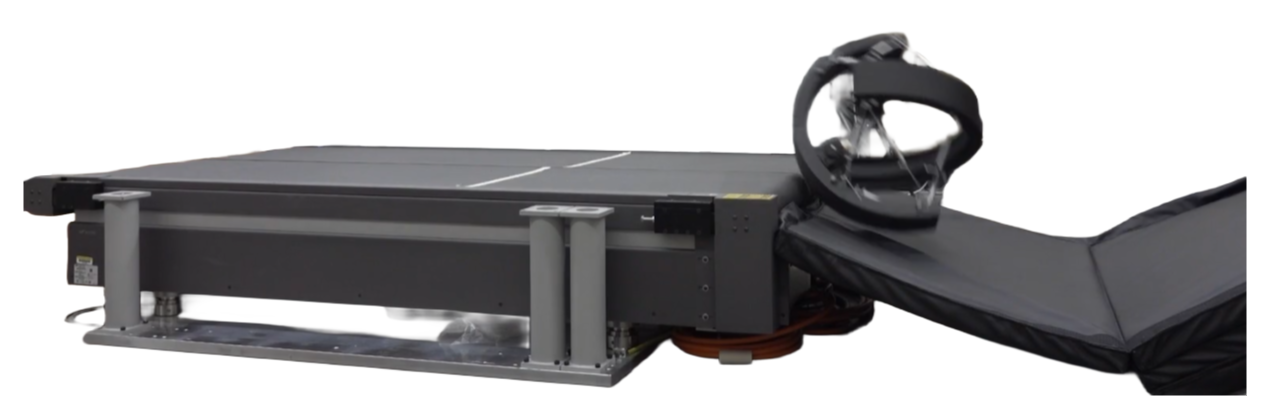}
         \caption{t=8.5}
         \label{fig:t85}
     \end{subfigure}
     
     \begin{subfigure}{0.49\linewidth}
         \centering
         \includegraphics[width=\textwidth]{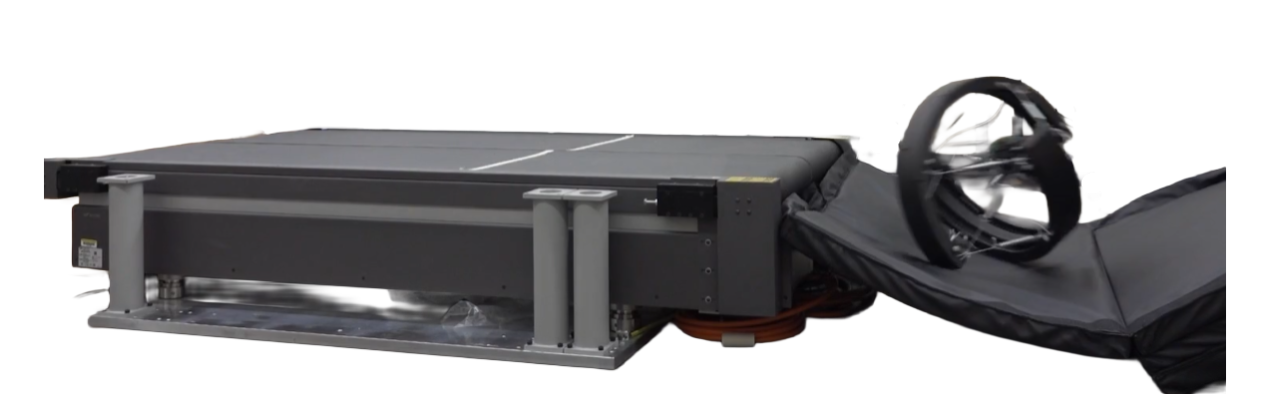}
         \caption{t=9}
         \label{fig:t9}
     \end{subfigure}
     \begin{subfigure}{0.49\linewidth}
         \centering
         \includegraphics[width=\textwidth]{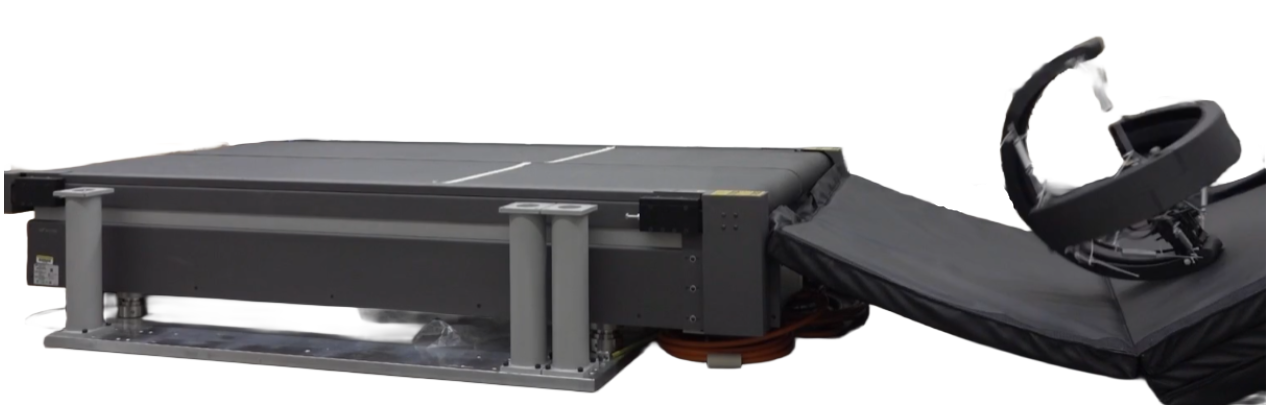}
         \caption{t=9.5}
         \label{fig:t95}
     \end{subfigure}
    \caption{A dynamic rolling sequence of TeXploR2 traversing along a stationary treadmill with a successful impact test at the end on a rubber mat.}
    \label{fig:impact}
\end{figure*}

\subsection{Real-world experiments}
The TeXploR2 prototype used for experiments shown above is also captured in the supplemental video.  In \Fig~\ref{fig:impact}, a dynamic rolling sequence traveling from state 1 $\rightarrow$ state 4 $\rightarrow$ state 2 is shown. Catching occurred on the internal gear racks that prevented the shifting masses from traveling the entirety of the arcs. This enabled a modified rolling sequence with non-intuitive switching behavior where state jumping occurred along a red boundary line. Instead of the typical rolling behavior where the shifting masses take turns traveling the entirety of their respective arc, non-intuitive jumping between states is exhibited. This shows jumping between states is achievable faster than is possible with the intuitive state switching discussed in the quasistatic analysis. It can also result in faster velocity; during the non-intuitive rolling, TeXploR2 rolled up to 1.88 BL/s.

Additional dynamic sequences other than rolling are also achievable. The specified rolling sequence produces movement in a straight trajectory. However, when traveling between waypoints, the trajectory of TeXploR2 can change with jumping motions, e.g., directly from state 1 $\rightarrow$ state 2 without entering state 4 as previously described. This could be done by moving one mass partially along its arc, stopping, and shifting the other mass from one endpoint to another. Depending on the position of the first mass, this will alter the trajectory of TeXploR2.

In \Fig~\ref{fig:impact}g and \Fig~\ref{fig:impact}h, TeXploR2 underwent an impact test. It hit a rubber mat head on that was at a roughly 600mm lower elevation than the stationary treadmill. During this impact, TeXploR2 experienced a slight deformation upon impact due to the compliant nature of the tensegrity primitive design. Critically, the arcs absorbed the shock without damaging any components.

\begin{figure*}[!ht]
    \vspace{0.5em}
    \centering
    \begin{subfigure}{0.46\linewidth}
         \centering
         \includegraphics[width=\textwidth]{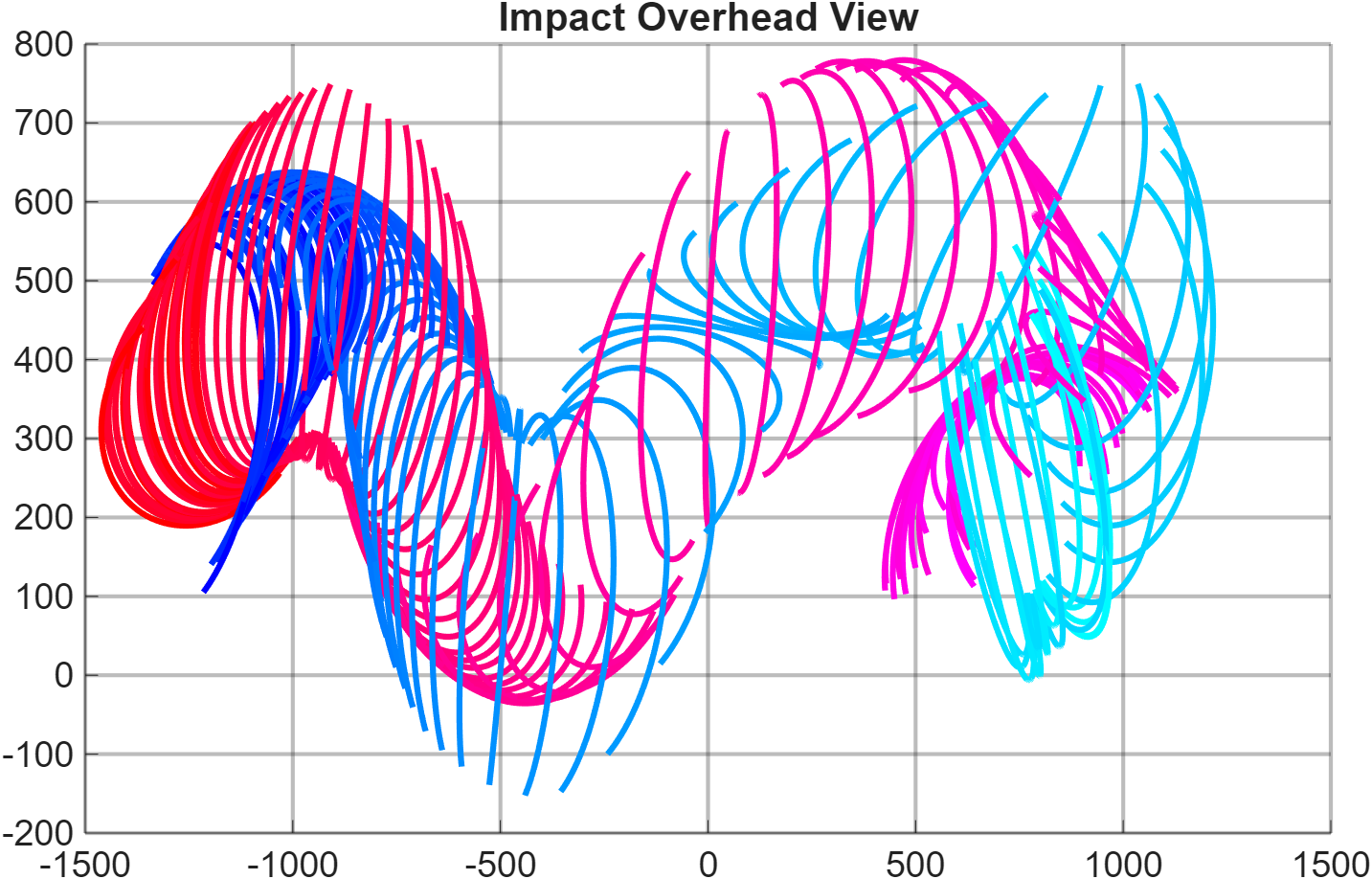}
         \caption{Overhead view}
         \label{fig:impact_over}
     \end{subfigure}
     \begin{subfigure}{0.46\linewidth}
         \centering
         \includegraphics[width=\textwidth]{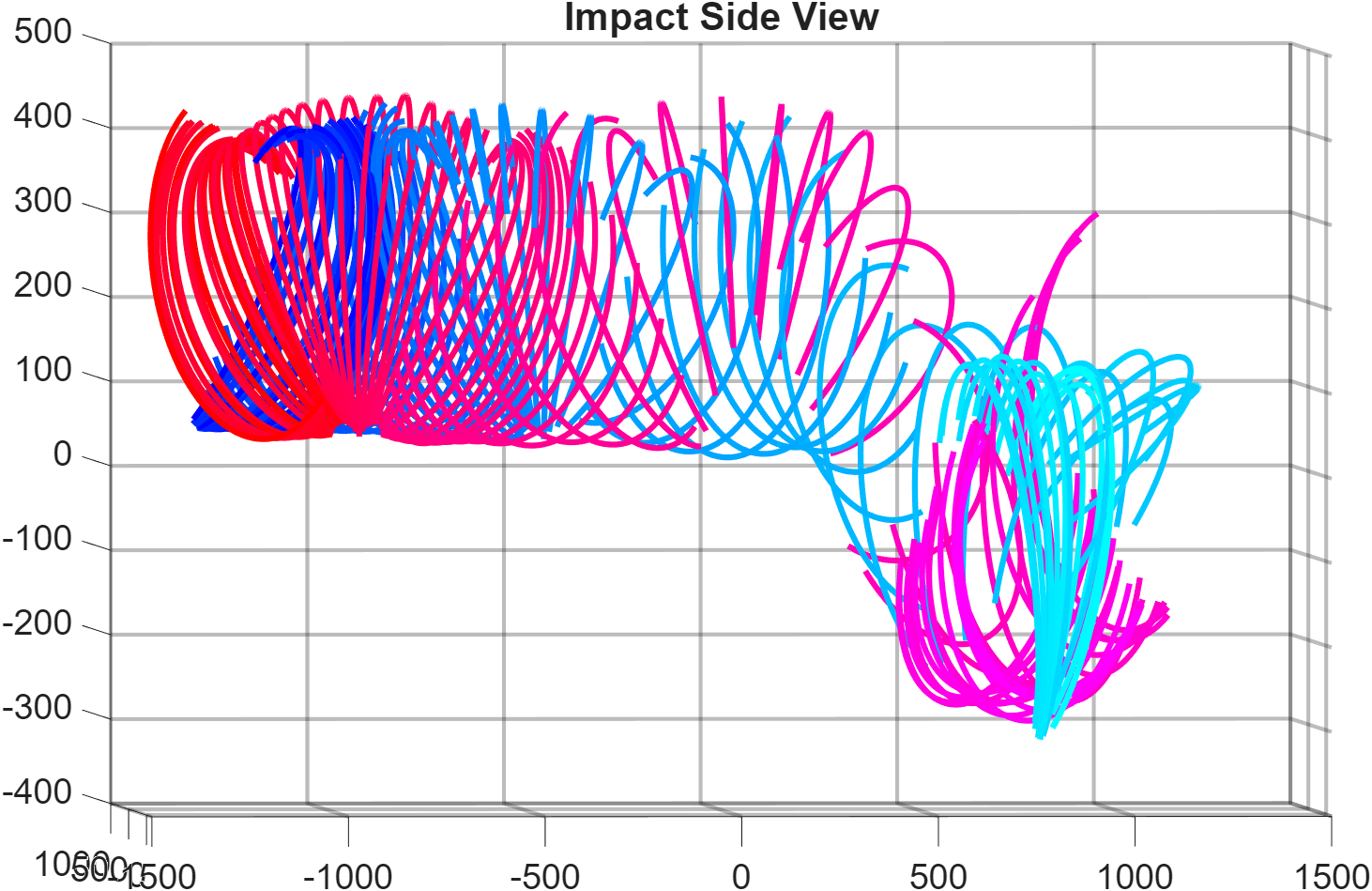}
         \caption{Side view}
         \label{fig:impact_side}
     \end{subfigure}
    \caption{Reconstructed points of the dynamic rolling sequence with TeXploR along a stationary treadmill with a successful impact test at the end on a rubber mat. The overhead and side views are provided to showcase the winding rolling as well as the impact test. The two arcs are represented by the red to magenta lines and blue to cyan lines.}
    \label{fig:impact_matlab}
\end{figure*}

\Fig~\ref{fig:impact_matlab} shows reconstructed points of the dynamic rolling sequence in \Fig~\ref{fig:impact} from a Vicon motion capturing system. The Vicon system is accurate up to 0.3mm with dynamic movement \cite{Vicon2020}. Since a test of general trajectory shape was the outcome of this dynamic test, this level of error could have gone up several orders of magnitude and not contributed heavily to the outcome of the experiment. The overhead view is provided to show that the rolling movement before the impact is similar to about half of the quasistatic simulation in \Fig~\ref{fig:fourstates} (state 1 $\rightarrow$ state 4). When a shifting mass caught on the internal gear rack after that point, initial rocking back and forth occurred before traveling in a modified path to the next state, state 2. Then, TeXploR2 entered a short period of free fall before impacting the lower rubber mat. The side view offers a better perspective of this drop. The impact test shows that TeXploR2 is capable of successfully withstanding changes in high acceleration without damage, and further testing in real-world environments is planned.

\section{CONCLUSION AND FUTURE WORK}
The system design for an improved tetherless TeXploR2 capable of dynamic movement sequences via internal mass shifting is detailed. MATLAB simulations of a four-state, quasistatic piecewise continuous rolling sequence are shown along with solved $x,y$ positions in the displacement vector $\bm{o}_{sb}$ that were previously thought to be free positions. A modified rolling sequence successfully exhibiting non-intuitive behavior is performed experimentally with the tetherless prototype. An impact test shows the resilience of the platform and highlights the importance of shock absorption capabilities of the compliant tensegrity primitive design. The results of a Vicon motion capturing system provide ground truth positioning to further analyze the experimental data. Future works include integrating an additional arc to make TeXploR2 a three point-of-contact hybrid system. This will increase both the controllability and maneuverability of the system which will be critical in constrained, highly tortuous environments that demand physical intelligence. However, to achieve this environmental adaptability, feedback control will be necessary. In these conditions, low sensing latency and path planning will play a large role in mission success. By scaling up and leaving a large open area in the center of the robot, a future direction of this research includes suspending a camera or LiDAR to increase the computational intelligence and sensing capabilities of TeXploR2.

\section*{ACKNOWLEDGMENT}
The authors thank Michael Faris for his thoughtful discussions and input on the TeXploR2 design.



\bibliographystyle{IEEEtran}
\bibliography{TeXploR.bib}
\end{document}